\pdfoutput=1
\PassOptionsToPackage{dvipsnames}{xcolor} 
\documentclass[11pt]{article}

\usepackage[]{acl}

\usepackage{times}
\usepackage{latexsym}
\usepackage[utf8]{inputenc}
\usepackage[T1]{fontenc}
\usepackage{CJKutf8}
\usepackage[english]{babel}
\usepackage{makecell}
\usepackage{graphicx}
\usepackage{hyperref}
\usepackage{multirow} 
\usepackage{float} 
\usepackage{amsmath} 
\usepackage{adjustbox} 
\usepackage{verbatim} 
\usepackage{hwemoji}
\usepackage{pifont} 
\usepackage[normalem]{ulem}
\useunder{\uline}{\ul}{}
\usepackage[dvipsnames]{xcolor} 

\usepackage[T1]{fontenc}

\usepackage[utf8]{inputenc}

\usepackage{microtype}

\title{\textsc{UniGen}: Universal Domain Generalization \\ for Sentiment Classification via Zero-shot Dataset Generation}

\author{Juhwan Choi\textsuperscript{1}, Yeonghwa Kim\textsuperscript{1}, Seunguk Yu\textsuperscript{1}, Jungmin Yun\textsuperscript{1} \and YoungBin Kim\textsuperscript{1,2} \\\\
  \textsuperscript{1}Department of Artificial Intelligence, Chung-Ang University \\
  \textsuperscript{2}Graduate School of Advanced Imaging Sciences, Multimedia and Film, Chung-Ang University \\
  \texttt{\{gold5230, movie112, bokju128, cocoro357, ybkim85\}@cau.ac.kr} \\
}

\begin{document}
\maketitle

\begin{abstract}
Although pre-trained language models have exhibited great flexibility and versatility with prompt-based few-shot learning, they suffer from the extensive parameter size and limited applicability for inference. Recent studies have suggested that PLMs be used as dataset generators and a tiny task-specific model be trained to achieve efficient inference. However, their applicability to various domains is limited because they tend to generate domain-specific datasets. In this work, we propose a novel approach to universal domain generalization that generates a dataset regardless of the target domain. This allows for generalization of the tiny task model to any domain that shares the label space, thus enhancing the real-world applicability of the dataset generation paradigm. Our experiments indicate that the proposed method accomplishes generalizability across various domains while using a parameter set that is orders of magnitude smaller than PLMs.
\end{abstract}

\section{Introduction}

\begin{table*}[]
\begin{center}
\resizebox{0.9\textwidth}{!}{
\begin{tabular}{c|cccc}
\Xhline{3\arrayrulewidth}
                                                                         & \begin{tabular}[c]{@{}c@{}}Learning without \\ Human-annotated Data\end{tabular} & \begin{tabular}[c]{@{}c@{}}Domain\\ Generalizability\end{tabular} & \begin{tabular}[c]{@{}c@{}}Light\\ Inference\end{tabular} & \begin{tabular}[c]{@{}c@{}}Handling Noise\\ of Generated Data\end{tabular} \\ \hline\hline
\begin{tabular}[c]{@{}c@{}}Task-specific Fine-tuning\end{tabular}     & \color{BrickRed}{\ding{55}}                                                                       & \color{BrickRed}{\ding{55}}                                                                & \color{ForestGreen}{\ding{51}}                                                        &                                                                        \\ \hline
\begin{tabular}[c]{@{}c@{}}Previous Domain Generalization \\ \cite{tan2022domain}\end{tabular} & \color{BrickRed}{\ding{55}}                                                                       & \color{ForestGreen}{\ding{51}}                                                                & \color{ForestGreen}{\ding{51}}                                                        &                                                                         \\ \hline
\textsc{Prompting}                                                        & \color{ForestGreen}{\ding{51}}                                                                       & \color{ForestGreen}{\ding{51}}                                                                & \color{BrickRed}{\ding{55}}                                                        &                                                                          \\ \hline
\textsc{ZeroGen} \cite{ye2022zerogen}                                                        & \color{ForestGreen}{\ding{51}}                                                                       & \color{BrickRed}{\ding{55}}                                                                & \color{ForestGreen}{\ding{51}}                                                        & \color{BrickRed}{\ding{55}}                                                                         \\ \hline
\begin{tabular}[c]{@{}c@{}}\textsc{ProGen} \& \textsc{SunGen} \\ \cite{ye2022progen, gao2023self}\end{tabular}                                      & \color{ForestGreen}{\ding{51}}                                                                       & \color{BrickRed}{\ding{55}}                                                                & \color{ForestGreen}{\ding{51}}                                                        & \color{ForestGreen}{\ding{51}}                                                                         \\ \hline
\textsc{UniGen} (Ours)                                                          & \color{ForestGreen}{\ding{51}}                                                                       & \color{ForestGreen}{\ding{51}}                                                                & \color{ForestGreen}{\ding{51}}                                                        & \textcolor{ForestGreen}{\ding{51}}                                                                         \\ \Xhline{3\arrayrulewidth}
\end{tabular}
}
\end{center}
\caption{Comparison between previous approaches and \textsc{UniGen}.}
\label{tab-comparison}
\end{table*}

As the size and performance of pre-trained language models (PLMs) increase, generation of new data by using PLMs has attracted the attention of many researchers \cite{anaby2020not, kumar2020data, yoo2021gpt3mix}. While scholars have applied this method to solve data augmentation problems, in recent studies, they have started to explore zero-shot dataset generation settings \cite{meng2022generating, ye2022zerogen, ye2023generating}. This novel approach first generates training data from a PLM based on a specific prompt and trains a tiny task model (TAM) by using the dataset generated in the first step. This strategy facilitates effective distillation of the knowledge pertaining to the desired task from the PLM and helps train the TAM without the need for guidance from human-annotated data, thereby enabling zero-shot learning and achieving low-cost inference compared to the case in which PLMs are used directly for inference.

However, the approaches proposed thus far have relied on domain-specific prompts, for example, ``\textit{The movie review in positive sentiment is:}.'' Because the data generated using this prompt are related only to the domain of movie reviews, the TAM trained on these data has limited generalization ability across other domains. This is the primary limitation of the TAM-based approach compared to prompt-based zero-shot learning that directly uses PLMs (\textsc{Prompting}), which allows for generalizability across diverse domains. This restricts the real-world applicability of the TAM-based approach because it requires many separately trained TAMs for various domains. Moreover, as the costs of dataset generation and TAM training increase, the cost-efficiency of the TAM-based approach may decrease. Hence, a novel strategy is desired to effectively distill the domain generalizability of large-scale PLMs into TAMs while maintaining the cost-efficiency of TAMs.

Meanwhile, the existing approaches to domain generalization often require multiple source domains \cite{wang2022generalizing, zhou2022domain}. This requirement limits the application of these methods because it is difficult to gather the required data from multiple domains. Although the concept of single-domain generalization, which achieves domain generalizability by using data from only one source domain, has been proposed in recent computer vision studies, such a concept is yet to be explored for natural language processing \cite{qiao2020learning, wang2021learning}.

In this study, we propose a simple but effective method called \textsc{UniGen} to solve the problem of domain generalizability between PLMs and TAMs. Table~\ref{tab-comparison} presents a comparison between \textsc{UniGen} and the existing approaches. \textsc{UniGen} first focuses on generating a domain-invariant training dataset that is not restricted to specific domains. This allows TAMs to achieve domain generalizability without the need for multiple source domains. We extend domain generalization strategies based on supervised contrastive learning \cite{khosla2020supervised}, as suggested in a previous work \cite{tan2022domain}. Moreover, we employ additional tactics such as momentum encoder \cite{he2020momentum} and denoised memory bank, in addition to the method suggested by the previous work \cite{tan2022domain}. Furthermore, because the PLM-based dataset generation method can generate noisy data \cite{ye2022progen, gao2023self, zou2024fusegen}, we propose a pseudo-relabeling-based additional denoising method.

Our experiments show that \textsc{UniGen} achieves generalizability across various domains and outperforms \textsc{Prompting}. This indicates that smaller TAMs can be used universally in various domains, thereby reducing the costs of \textsc{Prompting}, dataset generation, and TAM training.

Our contributions are summarized as follows: 

\begin{itemize}

\item We propose \textsc{UniGen}, a universal domain generalization strategy by using zero-shot dataset generation. 
\item We develop a pseudo-relabeling-based method for denoising the generated data. 
\item Our extensive experiment reveals that the TAM trained using \textsc{UniGen} has domain generalizability, and it can outperform the PLM with considerably fewer parameters.

\end{itemize}

\section{Related Work}

\subsection{Dataset Generation for Efficient Zero-shot Learning}
The evolution of PLMs in terms of parameter size and performance has facilitated zero-shot learning through the use of well-designed prompts \cite{radford2019language, brown2020language}. However, it is expensive to directly deploy these massive models into daily services because the process requires numerous rounds of inference. Dataset generation mitigates this problem through the generation of training datasets by using PLMs and training a small TAM on the generated datasets \cite{meng2022generating, ye2022zerogen}. This TAM is deployed in downstream tasks to reduce inference costs and improve performance compared to \textsc{Prompting}.

However, mere generation, that is, \textsc{ZeroGen}, yields noisy data, such as incorrectly labeled data or irrelevant data \cite{ye2022progen, gao2023self}. \textsc{ProGen} \cite{ye2022progen} proposed to alleviate this problem by adding examples based on in-context feedback. Meanwhile, \textsc{SunGen} \cite{gao2023self} proposed to re-weigh the generated samples during training using noise-robust loss. Additionally, a concurrent study suggested to leverage multiple PLMs as data generator and assign weight to generated samples in single training procedure, different from \textsc{SunGen} \cite{zou2024fusegen}.

In this work, we propose a novel approach to extend dataset generation for universal domain generalization that is not restricted to specific training source data, as well as a pseudo-relabeling-based method to denoise the generated dataset.

\subsection{Methods for Learning from Noisy Data}
Researchers have explored various methods to mitigate noisy label data, which is wrongly labeled from ground-truth labels \cite{song2022learning}. A relevant study in this field defined two types of noisy labels and evaluated the effectiveness of various methods with respect to BERT model \cite{agro2023handling}. Another study proposed to leverage GPT-4 to provide the guidance to noisy labeled data \cite{wang2023noise}. However, they suffer from the necessity of massive LLMs that demand cost. Moreover, these studies primarily focused on the human-crafted noisy label, rather than the noisy label of data generated by PLMs.

In this work, we suggest a straightforward method to handle noisy data based on pseudo-relabeling, particularly designed for synthetic data.

\subsection{Domain Generalization for Text Classification}
Domain generalization aims to improve the generalization ability in the target domain by employing source data from multiple domains to mitigate the domain shift problem \cite{wang2022generalizing, zhou2022domain}. This domain shift can be observed in natural language processing tasks, such as restaurant reviews and reviews of consumer electronics. For example, \textit{long waiting time} in a restaurant’s reviews can represent a negative sentiment about the restaurant, while \textit{long battery life} in a laptop’s reviews can represent a positive sentiment of the laptop \cite{tan2022domain}.

Previous studies to alleviate domain shift in text classification have focused primarily on domain adaptation setting, for which training data are needed in the target domain \cite{chen2018multinomial, ye2020feature, guo2020multi}. Recently, researchers have explored the application of domain generalization to natural language processing tasks. A representative study applied supervised contrastive learning \cite{khosla2020supervised} to achieve domain generalizability in text classification tasks \cite{tan2022domain}.

In this work, we extend an existing method for domain generalization to generate datasets, including the adoption of momentum encoder \cite{he2020momentum}, in addition to proposing a denoising memory bank to further enhance its effectiveness and handle noisy data.

\section{Method}

\begin{figure*}[t]
    \centering
    \includegraphics[width=\textwidth]{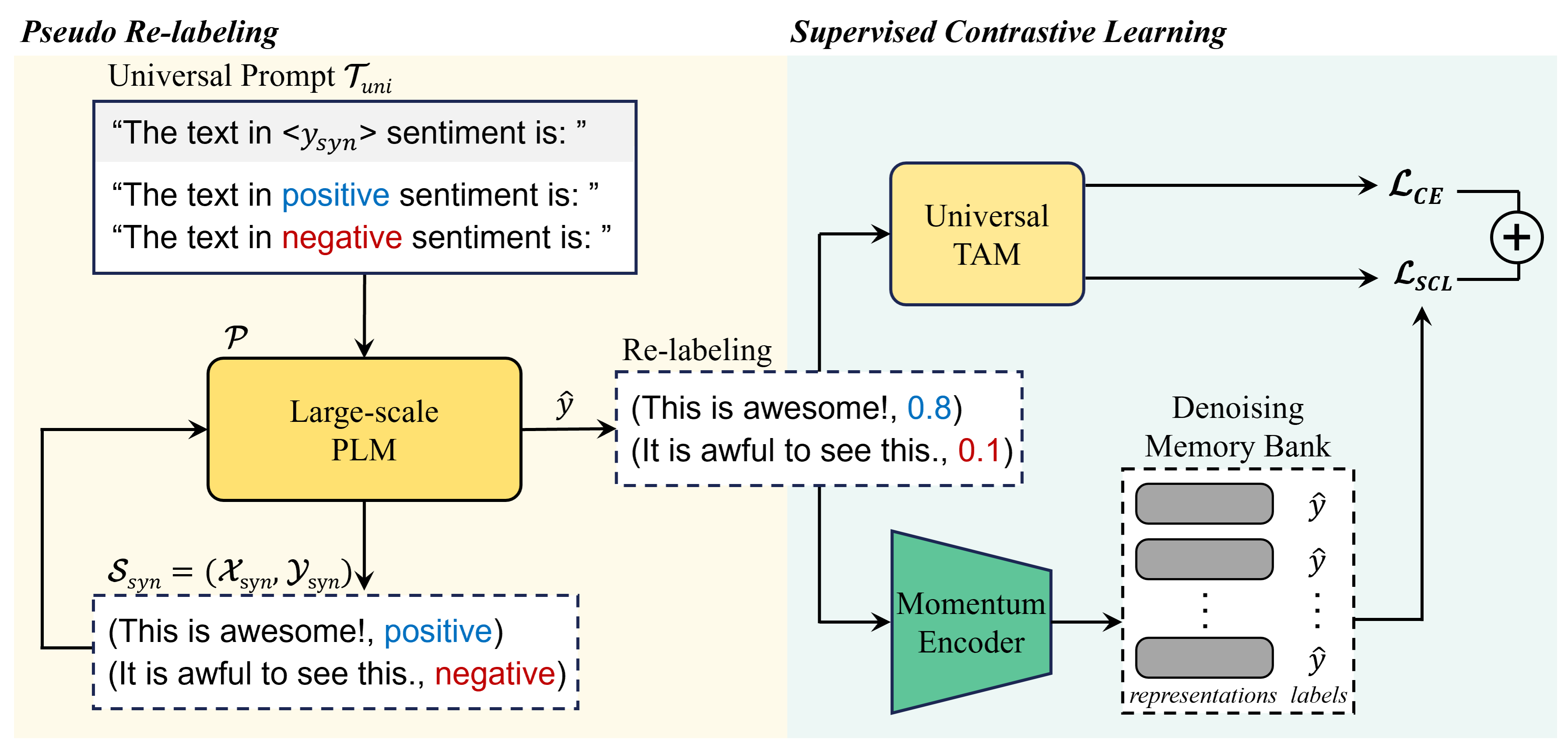}
    \caption{Overall framework for generating a dataset and training a TAM using \textsc{UniGen}.}
\label{fig-framework}
\end{figure*}

\subsection{Preliminaries}
\subsubsection{Dataset Generation}
First, we briefly explain the concept and notation of the preliminary dataset generation method, that is, \textsc{ZeroGen} \cite{ye2022zerogen}. \textsc{ZeroGen} aims to create a synthetic dataset $\mathcal{S}_{\textit{syn}} = (\mathcal{X}_{\textit{syn}}, \mathcal{Y}_{\textit{syn}})$ by using a large-scale PLM $\mathcal{P}$ and task-specific prompt $\mathcal{T}_{\textit{task}}$. For a text classification problem, a desired pseudo-label $y_{\textit{syn}}$ is first sampled from the uniform distribution across every class. Next, $y_{\textit{syn}}$ is passed to the prompt $\mathcal{T}_{\textit{task}}$ to construct $\mathcal{T}_{\textit{task}}(y_{\textit{syn}})$, that is, the final prompt for $\mathcal{P}$. Thereafter, synthesized input data $x_{syn}$ are generated using $\mathbf{x}_{\textit{syn}} \sim \mathcal{P}(\cdot|\mathcal{T}_{\textit{task}}(y_{\textit{syn}}))$. Finally, $\mathcal{S}_{\textit{syn}}$ is composed of these pairs of generated $(\mathbf{x}_{\textit{syn}}, y_{\textit{syn}})$. Notably, the domain of $\mathcal{S}_{\textit{syn}}$ is defined by the structure of $\mathcal{T}_{\textit{task}}$. For example, a $\mathcal{T}_{\textit{book}} = $ ``\textit{The book review in <$y$> sentiment is: }'' would harness $ \mathcal{P}$ to generate $\mathbf{x}_{\textit{syn}}$ about book reviews. The TAM is trained on the generated $\mathcal{S}_{\textit{syn}}$ and deployed for inference instead of directly using PLMs with \textsc{Prompting}.

\subsubsection{Supervised Contrastive Learning}
\label{sec:method-supcon}

Supervised contrastive learning \cite{khosla2020supervised} is a variant of contrastive learning \cite {chen2020simple} that utilizes label values. It allows for explicit pulling of the representation of positive (i.e., same class) samples to the anchor representation while pushing negative representations away from the anchor. Studies have reported that this characteristic is valuable for domain generalization, which aims to group the representations of different domains \cite{kim2021selfreg, tan2022domain}. The supervised contrastive loss is expressed as follows:

\begin{equation}
\begin{aligned}
\resizebox{\columnwidth}{!}{
$\mathcal{L}_{\textit{SCL}} = -\sum_{\mathbf{z}_i \in B} \frac{1}{|P(i)|} log \frac{exp(\mathbf{z}_i \cdot \mathbf{z}_p / \tau_{\textit{SCL}})}{\sum_{\mathbf{z}_a \in A(i)} exp(\mathbf{z}_i \cdot \mathbf{z}_a / \tau_{\textit{SCL}})}$
}
\end{aligned}
\end{equation}

where $\mathbf{z}$ denotes an encoded representation, and $\mathbf{z_i}$ is an anchor. $P(i) \equiv {\mathbf{z}_j \in B, y_j = y_i}$ is the set of positive samples for each anchor $i$, and $\mathbf{z}_p$ symbolizes a positive representation from $P(i)$. $A(i) \equiv {\mathbf{z}_j \in B, j \neq i}$ refers to the union of every sample, except the anchor, including positive and negative samples. $\mathbf{z}_a$ indicates each representation from $A(i)$. $B$ denotes a mini-batch, and $\tau_{\textit{SCL}}$ is the temperature of supervised contrastive learning.

Although supervised contrastive learning is effective, the introduction of a memory bank and momentum encoder may augment the advantages of the method \cite{wu2018unsupervised, he2020momentum}. The potency of contrastive learning is often influenced by the size of $B$ because a larger $B$ may introduce more diverse negative samples. However, increasing the size of $B$ can introduce concerns related to memory consumption. A memory bank is a mechanism that fulfills this demand for a greater number of negative samples by storing previously processed samples within the dictionary $M$. Memory-efficient contrastive learning can be achieved using this dictionary with the current batch, that is, establishing a union of $B$ and $M$ instead of solely using $B$ to construct $P(i)$ and $A(i)$. Momentum encoder is another technique that smooths the process of updating the representations stored in $M$. The momentum encoder $\theta_{k}$ is trained by momentum update, $\theta_{k} \gets m \theta_{k} + (1-m)\theta_{q}$, where $m$ is a coefficient for momentum update, and $\theta_{q}$ is a normal encoder that is updated through backpropagation. By using the momentum encoder, the representations in $M$ are processed by $\theta_{k}$.

\subsection{\textsc{UniGen}}

To build a TAM that can be applied universally to various target domains, \textsc{UniGen} generates a domain-invariant dataset by using the universal prompt $\mathcal{T}_{\textit{uni}}$, instead of task-specific $\mathcal{T}_{\textit{task}}$. Consider ``\textit{The text in <$y$> sentiment is:}'' as an example of $\mathcal{T}_{\textit{uni}}$. Next, the final input prompt for $\mathcal{P}$ is constructed as $\mathcal{T}_{\textit{uni}}(y_{\textit{syn}})$. The synthesized input data $x_{\textit{syn}}$ are generated by following the same process as that of \textsc{ZeroGen}:

\begin{equation}
\begin{aligned}
\mathbf{x}_{\textit{syn}} \sim \mathcal{P}(\cdot|\mathcal{T}_{\textit{uni}}(y_{\textit{syn}}))
\end{aligned}
\end{equation}

This configuration of prompt design allows us to generate a sentence with the desired label without being restricted to any specific domain. Therefore, it steers $\mathcal{P}$ to generate various sentences within a predefined label space. This domain-invariant data generation allows the TAM trained using \textsc{UniGen} to learn the domain-invariant characteristics of the desired label space, thereby resulting in generalizability across the domains that share the label space. Supervised contrastive loss is applied along with conventional cross entropy loss to aid this process. The training loss is defined as follows:

\begin{equation}
\begin{aligned}
\mathcal{L} = \mathcal{L}_{\textit{CE}} + \alpha\mathcal{L}_{\textit{SCL}}
\end{aligned}
\end{equation}

where $\alpha$ is a hyperparameter that balances the ratio between the two losses.

\subsection{Handling Noisy Data through Relabeling}
\label{sec:method-relabeling}

However, the application of $\mathcal{T}_{\textit{uni}}$ instead of $\mathcal{T}_{\textit{task}}$ might lead to the generation of noisy sentences, which was noted as a drawback of \textsc{ZeroGen}. This is because $\mathcal{T}_{\textit{uni}}$ does not have a specific topic to guide the generation process. Furthermore, a previously developed approach to effectively mitigate this problem is applied in the training phase but not the generation phase. Therefore, there is scope to improve the quality of $\mathcal{S}_{\textit{syn}}$ \cite{gao2023self}. This problem highlights the necessity to use a denoising scheme in the generation procedure. In the present work, we propose a pseudo-relabeling-based denoising process for dataset generation. In a previous study, the approach of relabeling the generated data and assigning soft labels for data augmentation was proposed \cite{yoo2021gpt3mix}. Herein, we first perform pseudo-relabeling by using $\mathcal{P}$:

\begin{equation}
\begin{aligned}
\ell(y_i|\mathbf{x}_{\textit{syn}}) = \mathcal{P}(\mathcal{M}(y_i)|\mathcal{T}_{\textit{uni}}(\mathbf{x}_{\textit{syn}}))
\end{aligned}
\end{equation}

where $\mathcal{M}(\cdot)$ denotes a verbalizer that transforms each label $y_i$ into a word. We share $\mathcal{T}_{\textit{uni}}$ between this process and the generation process. These logit values yielded by $\mathcal{P}$ are normalized using the softmax function with the temperature $\tau_{\textit{RE}}$ :

\begin{equation}
\begin{aligned}
\hat{y_i} = p(y_i|\mathbf{x}_{\textit{syn}}) = \frac{exp(\ell(y_i|\mathbf{x}_{\textit{syn}}) / \tau_{\textit{RE}})}{\sum_{j} exp(\ell(y_j|\mathbf{x}_{\textit{syn}}) / \tau_{\textit{RE}})}
\end{aligned}
\end{equation}

Finally, we assign $\hat{y_i}$ instead of the predefined $y_{\textit{syn}}$ to the generated $\mathbf{x}_{\textit{syn}}$. This provides two distinct advantages: (1) because $\hat{y_i}$ is a soft label rather than a hard label, it contains richer information about $\mathbf{x}_{\textit{syn}}$, such as the degree of the desired label, which enhances the effectiveness of training \cite{szegedy2016rethinking}. (2) Because it relabels the generated $\mathbf{x}_{\textit{syn}}$ and replaces the predefined $y_{\textit{syn}}$, it can solve the noisy label issue, which results in the generation of $\mathbf{x}_{\textit{syn}}$ that does not correspond to the designated $y_{\textit{syn}}$, as pointed out in previous work \cite{gao2023self}. We validate the effectiveness of this relabeling strategy in the ablation study described in Section~\ref{exp:ablation1}.

Furthermore, we discard $\mathbf{x}_{\textit{syn}}$ if its pseudo-label $\hat{y_i}$ does not exceed the threshold $T_{\textit{RE}}$ to enhance the quality of $\mathcal{S}_{\textit{syn}}$. This guarantees that only those data that have the desired degree of each label are maintained.

\subsection{Denoising Memory Bank}
\label{sec:method-denoisingMB}
In addition to the relabeling strategy, we propose a denoising memory bank mechanism to further alleviate the issue of noisy data. We first use \textsc{SunGen} \cite{gao2023self} that learns weights of each training sample $\mathbf{w}$ for loss function within the training process to assign small weights to noisy data by employing a noise-robust loss function. We aim to ensure that the memory bank $M$ contains clean samples, rather than noisy samples. We utilize the weights $\mathbf{w}$ learned from the noise-robust loss function for this purpose. In the process of updating $M$, we store only those samples whose weights are larger than the threshold $T_{\textit{MB}}$. This organization of the memory bank ensures the exclusion of noisy samples from the comparison, resulting in higher-quality negative and positive samples \cite{robinson2020contrastive}.

\section{Experiment}


\begin{table*}[t]
\begin{center}
\resizebox{\textwidth}{!}{
\begin{tabular}{cc|cc|ccccccc|c}
\Xhline{3\arrayrulewidth}
Model                        & \#Param                & Training Domain             & Setup              & SST-2                & IMDB                 & Rotten               & Amazon               & Yelp                 & CR                   & Tweet                & Average              \\
Test Domain                  &                        &                             &                    & \multicolumn{3}{c}{Movie}                                          & Products             & Restaurant           & Electronics          & Tweet                &                      \\ \hline\hline
GPT2-XL                      & 1.5B                   & -                           & \textsc{Prompting} & 82.15                & 70.26                & 77.56                & 79.06                & 78.04                & 80.30                & 80.38                & 78.25                \\ \Xhline{2\arrayrulewidth}
\multirow{11}{*}{LSTM}       & \multirow{11}{*}{7M}   & \multirow{2}{*}{Movie}      & \textsc{ZeroGen}   & 75.11                & 66.39                & 69.85                & 67.24                & 70.25                & 69.32                & 63.43                & 68.80                \\
                             &                        &                             & \textsc{SunGen}    & \textbf{78.79}       & \textbf{69.97}       & \textbf{73.76}       & 72.15                & 73.21                & 70.39                & 66.84                & \textbf{72.16}       \\ \cline{3-12} 
                             &                        & \multirow{2}{*}{Products}   & \textsc{ZeroGen}   & 64.26                & 61.82                & 60.13                & 70.32                & 67.78                & 69.46                & 62.29                & 65.15                \\
                             &                        &                             & \textsc{SunGen}    & 67.83                & 63.87                & 63.46                & \textbf{74.43}       & 73.71                & 73.35                & 63.51                & 68.59                \\ \cline{3-12} 
                             &                        & \multirow{2}{*}{Restaurant} & \textsc{ZeroGen}   & 67.41                & 63.01                & 62.74                & 68.73                & 75.51                & 69.23                & 66.35                & 63.28                \\
                             &                        &                             & \textsc{SunGen}    & 69.15                & 66.62                & 64.56                & 73.22                & {\ul \textbf{79.56}} & 70.12                & 67.43                & 70.09                \\ \cline{3-12} 
                             &                        & \multirow{2}{*}{Electronics}& \textsc{ZeroGen}   & 64.69                & 59.13                & 60.20                & 66.34                & 67.72                & 72.50                & 60.25                & 64.40                \\
                             &                        &                             & \textsc{SunGen}    & 68.38                & 64.33                & 63.25                & 72.61                & 73.01                & \textbf{76.18}       & 66.78                & 69.22                \\ \cline{3-12} 
                             &                        & \multirow{2}{*}{Tweet}      & \textsc{ZeroGen}   & 61.84                & 60.17                & 59.43                & 64.13                & 63.68                & 65.02                & 74.10                & 64.05                \\
                             &                        &                             & \textsc{SunGen}    & 66.57                & 63.96                & 64.21                & 69.36                & 71.68                & 72.57                & {\ul \textbf{81.29}} & 69.95                \\ \cline{3-12} 
                             &                        & -                           & \textsc{UniGen}    & 64.15                & 60.02                & 60.51                & 63.82                & 63.20                & 69.61                & 70.32                & 64.52                \\  \Xhline{2\arrayrulewidth}
\multirow{11}{*}{DistilBERT} & \multirow{11}{*}{66M}  & \multirow{2}{*}{Movie}      & \textsc{ZeroGen}   & 80.06                & 69.13                & 74.73                & 73.02                & 72.77                & 73.59                & 74.83                & 74.02                \\
                             &                        &                             & \textsc{SunGen}    & {\ul \textbf{82.43}} & {\ul \textbf{70.59}} & \textbf{76.37}       & 74.13                & 73.56                & 75.14                & 75.96                & 75.45                \\ \cline{3-12} 
                             &                        & \multirow{2}{*}{Products}   & \textsc{ZeroGen}   & 71.04                & 64.99                & 65.57                & 74.54                & 71.89                & 74.57                & 71.93                & 70.65                \\
                             &                        &                             & \textsc{SunGen}    & 72.35                & 65.95                & 66.84                & \textbf{76.92}       & 74.98                & 75.84                & 73.01                & 72.27                \\ \cline{3-12} 
                             &                        & \multirow{2}{*}{Restaurant} & \textsc{ZeroGen}   & 77.32                & 65.47                & 68.86                & 74.01                & 77.94                & 74.89                & 73.74                & 73.18                \\
                             &                        &                             & \textsc{SunGen}    & 78.93                & 67.12                & 69.92                & 74.93                & {\ul \textbf{80.67}} & 76.06                & 75.28                & 74.70                \\ \cline{3-12} 
                             &                        & \multirow{2}{*}{Electronics}& \textsc{ZeroGen}   & 73.77                & 66.14                & 66.78                & 72.38                & 73.21                & 78.82                & 74.58                & 72.24                \\
                             &                        &                             & \textsc{SunGen}    & 74.49                & 67.19                & 68.29                & 73.49                & 75.34                & {\ul \textbf{80.49}} & 75.37                & 73.52                \\ \cline{3-12} 
                             &                        & \multirow{2}{*}{Tweet}      & \textsc{ZeroGen}   & 73.98                & 66.58                & 67.43                & 72.88                & 71.86                & 75.68                & {\ul 80.86}          & 72.75                \\
                             &                        &                             & \textsc{SunGen}    & 75.12                & 67.53                & 69.06                & 73.64                & 72.73                & 78.17                & {\ul \textbf{82.46}} & 74.10                \\ \cline{3-12} 
                             &                        & -                           & \textsc{UniGen}    & 77.67                & 67.81                & 73.16                & 75.06                & 74.81                & 79.86                & {\ul 81.41}          & \textbf{75.68}       \\ \Xhline{2\arrayrulewidth}
\multirow{11}{*}{RoBERTa}    & \multirow{11}{*}{110M} & \multirow{2}{*}{Movie}      & \textsc{ZeroGen}   & {\ul 84.38}          & {\ul 73.03}          & {\ul 78.38}          & 77.38                & 76.83                & 77.36                & 77.94                & 77.90                \\
                             &                        &                             & \textsc{SunGen}    & {\ul \textbf{85.24}} & {\ul \textbf{74.09}} & {\ul \textbf{79.19}} & 78.56                & 77.61                & 78.21                & 79.72                & {\ul 78.95}                \\ \cline{3-12} 
                             &                        & \multirow{2}{*}{Products}   & \textsc{ZeroGen}   & 79.14                & {\ul 71.16}          & 70.92                & {\ul 79.94}          & 75.79                & 76.35                & 80.17                & 76.21                \\
                             &                        &                             & \textsc{SunGen}    & 81.51                & {\ul 71.28}          & 72.67                & {\ul \textbf{81.50}} & 77.76                & 78.55                & {\ul 81.94}                & 77.87                \\ \cline{3-12} 
                             &                        & \multirow{2}{*}{Restaurant} & \textsc{ZeroGen}   & {\ul 82.87}          & {\ul 70.71}          & 69.58                & 78.61                & {\ul 81.47}          & 76.43                & 79.51                & 77.03                \\
                             &                        &                             & \textsc{SunGen}    & {\ul 83.65}          & {\ul 71.40}          & 71.05                & {\ul 79.42}          & {\ul \textbf{82.72}} & 77.60                & {\ul 80.92}          & 78.11                \\ \cline{3-12} 
                             &                        & \multirow{2}{*}{Electronics}& \textsc{ZeroGen}   & 76.82                & 69.42                & 67.89                & 75.02                & 76.53                & {\ul 81.24}          & 76.51                & 74.78                \\
                             &                        &                             & \textsc{SunGen}    & 77.51                & {\ul 71.23}          & 68.77                & 76.91                & {\ul 78.33}          & {\ul 83.49} & 79.03                & 76.47              \\ \cline{3-12} 
                             &                        & \multirow{2}{*}{Tweet}      & \textsc{ZeroGen}   & 78.43                & 68.31                & 72.25                & 78.09                & 74.61                & 79.08                & {\ul 82.96}          & 76.25                \\
                             &                        &                             & \textsc{SunGen}    & {\ul 82.19}          & {\ul 70.62}          & 73.21                & {\ul 79.84}          & 76.27                & {\ul 81.46}          & {\ul 83.25}          & 78.12                \\ \cline{3-12} 
                             &                        & -                           & \textsc{UniGen}    & {\ul 84.86}          & {\ul 72.24}          & {\ul 78.82}          & {\ul 80.79}          & {\ul 79.15}          & {\ul \textbf{86.37}}          & {\ul \textbf{87.89}}          & {\ul \textbf{81.45}} \\ \Xhline{3\arrayrulewidth}
\end{tabular}
}
\end{center}
\caption{Experimental results of \textsc{UniGen} and baselines across various datasets and training domains. The performance of TAM, which is superior to that of \textsc{Prompting}, is underlined, and the best result in each test dataset within the group for each TAM is presented in boldface.}
\label{tab-main}
\end{table*}

\begin{table}[t]
\begin{center}
\resizebox{\columnwidth}{!}{
\begin{tabular}{c|cc}
\Xhline{3\arrayrulewidth}
                 & Amount of generated data & Number of trained TAMs \\ \hline\hline
\textsc{ZeroGen} & 1,000k                   & 5                  \\
\textsc{SunGen}  & 5,000k                   & 5                  \\
\textsc{UniGen}  & 1,000k                   & 1                  \\ \Xhline{3\arrayrulewidth}
\end{tabular}
}
\end{center}
\caption{Amount of data generated for training TAMs by using each method, and number of trained TAMs per method.}
\label{tab-amount}
\end{table}

\subsection{Experimental Setup}
In this section, we briefly explain the experimental setup used herein to validate the effectiveness of \textsc{UniGen}.
We employ seven different sentiment classification datasets in our main experiment. Among them, IMDB \cite{maas2011learning}, SST-2 \cite{socher2013recursive}, and Rotten Tomatoes \cite{pang2005seeing} are  datasets comprising movie reviews. Meanwhile, the Amazon \cite{mcauley2013hidden} dataset consists of customer reviews of various products, and the Yelp \cite{zhang2015character} dataset is composed of restaurant reviews. CR \cite{ding2008holistic} is another customer review dataset focusing on consumer electronics. Lastly, Tweet \cite{rosenthal2017semeval} is composed of messages from Twitter. This configuration allows us to evaluate the ability of \textsc{UniGen}, which can be applied to various domains without providing any prior information or domain-specific training. Following the previous study, we adapted long short-term memory (LSTM) \cite{hochreiter1997long} and DistilBERT \cite{sanh2019distilbert}, and we included RoBERTa \cite{liu2019roberta} as our TAM. We compared our approach to \textsc{ZeroGen} and \textsc{SunGen}, as well as to \textsc{Prompting} using GPT2-XL \cite{radford2019language}, to ensure a fair comparison. We did not include other larger PLMs in the experiments because the previous work discovered that larger PLMs did not offer performance gains \cite{ye2022zerogen}. We report the average of the performance results obtained across five different random seeds.

\subsection{Comparison with Task-specific TAMs}

Table~\ref{tab-main} presents a comparison between the experimental results of \textsc{UniGen} and \textsc{Prompting} and task-specific TAMs trained by \textsc{ZeroGen} and \textsc{SunGen}. The comparison results suggest that \textsc{UniGen} can generalize across various domains using a \textit{single} model \textit{without} requiring any prior information about the test domain. Nonetheless, \textsc{UniGen} underperformed compared to the task-specific baselines in each domain. However, the primary benefit of \textsc{UniGen} lies in its unique domain generalizability while using orders-of-magnitude fewer parameters than PLMs. Additionally, its training procedure is more efficient than those of other TAM training strategies. As can be inferred from Table~\ref{tab-amount}, \textsc{SunGen} generates and synthesizes 1,000k data for each task domain. This means that 5,000k data would be required for our experiment, which involves five different domains, in addition to individual denoising processes for finding the best weights of the samples in each of these domains. By contrast, \textsc{UniGen} is not limited by such restrictions and requires only a single data generation and denoising process, as well as a single training process. This is extremely beneficial when a novel test domain is introduced, where \textsc{ZeroGen} and \textsc{SunGen} necessitate a separate procedure for the new domain, but \textsc{UniGen} directly reuses the already trained TAM.

Notably, the performance of the LSTM-based TAM trained using \textsc{UniGen} was significantly lower than that of \textsc{ZeroGen} and \textsc{SunGen}. This implies that while a small-sized TAM can be trained effectively for a single, specific domain, but suffers from generalizing to a universal domain that requires a broad understanding of generated data, as evidenced by detailed study in Appendix~\ref{app:small-tam}. Accordingly, the performance of the TAM trained using \textsc{UniGen} improves significantly as the model size increases. For instance, the DistilBERT-based TAM trained using \textsc{UniGen} exhibited the best average performance against each task-specific baseline. This is particularly remarkable as it outperformed the \textsc{SunGen} baseline in the movie domain, which has three in-domain datasets, giving it an inherent advantage for average performance. Moreover, the RoBERTa-based TAM trained using \textsc{UniGen} not only yielded the best average performance against these baselines but also outperformed \textsc{Prompting} in every domain. This result indicates that it can surpass the zero-shot performance of its PLM counterpart (e.g., GPT2-XL) while using less than 10\% of the number of parameters and securing the domain generalizability of the PLM, extending the achievement of the previous study that leveraged small TAMs in single domain \cite{ye2022zerogen}.

\subsection{Comparison with Supervised Domain Generalization Method}

\begin{table}[]
\begin{center}
\resizebox{\columnwidth}{!}{
\begin{tabular}{c|ccccc}
\Xhline{3\arrayrulewidth}
RoBERTa                                                                            & DVD   & Electronics & Kitchen & Book  & Average \\ \hline\hline
\begin{tabular}[c]{@{}c@{}}\textsc{Prompting}\\ w/ GPT2-XL\end{tabular}            & 77.73 & 78.71       & 81.64   & 80.27 & 79.59 \\
\textsc{UniGen}                                                                    & 78.14 & 80.68       & 82.31   & 80.93 & 80.52 \\
\begin{tabular}[c]{@{}c@{}}\textsc{Supervised}\\ \cite{tan2022domain}\end{tabular} & 91.40 & 95.10       & 95.05   & 93.25 & 93.70 \\ \Xhline{3\arrayrulewidth}
\end{tabular}
}
\end{center}
\caption{Experiments conducted using multi-domain review dataset. The experimental result of \textsc{Supervised} was reported in a previous study \cite{tan2022domain} with the memory bank size of 64.}
\label{tab-supervised}
\end{table}

Next, we analyzed the performance of \textsc{UniGen} against that of a domain generalization method that uses human-annotated data \cite{tan2022domain}. For this purpose, we used a multi-domain review dataset comprising four domains: DVD, books, kitchen and housewares, and consumer electronics \cite{blitzer2007biographies}. Following the previous study, we split the dataset into 1,600 training data and 400 testing data for each domain. Table~\ref{tab-supervised} presents the comparison results. These results suggest that \textsc{UniGen} can be applied to various domains, and its performance is superior to that of its PLM counterpart. Notably, the \textsc{Supervised} baseline relies on three source domains with human-annotated data to generalize to a target domain, while \textsc{UniGen} is based on zero-shot dataset generation and does not require any human-annotated data, which greatly improves its real-world applicability.

\subsection{Domain Generalizability of \textsc{UniGen}}

\begin{table*}[t]
\begin{center}
\resizebox{\textwidth}{!}{
\begin{tabular}{ll}
\Xhline{3\arrayrulewidth}
Positive Examples                                                                                                                  & Labels         \\ \hline\hline
You are a person who is hardworking, honest, and reliable. You have a good sense of humor, and you love being in charge.           & $[0.19, 0.81]$ \\
You are beautiful, you are powerful, you are amazing.                                                                              & $[0.29, 0.71]$ \\
In a city full of great ideas and creativity, I've met a few people who have done things you wouldn't believe.                     & $[0.26, 0.74]$ \\
The American Dream is alive in this great city. As a new generation of American heroes begins to realize their own American Dream. & $[0.24, 0.76]$ \\ \Xhline{2\arrayrulewidth}
Negative Examples                                                                                                                  & Labels         \\ \hline\hline
No one likes it. Nobody wants it. It is a disgrace.                                                                                & $[0.7, 0.3]$   \\
The company is no longer in business and has ceased operations.                                                                    & $[0.71, 0.29]$ \\
Please don't use this feature to communicate with customers                                                                        & $[0.74, 0.26]$ \\
Do not buy from this seller.                                                                                                       & $[0.79, 0.21]$ \\ \Xhline{3\arrayrulewidth}
\end{tabular}
}
\end{center}
\caption{Examples of the data generated using \textsc{UniGen}.}
\label{tab-examples}
\end{table*}

\begin{figure}[h]
\vspace{-5mm}
    \centering
    \includegraphics[width=\columnwidth]{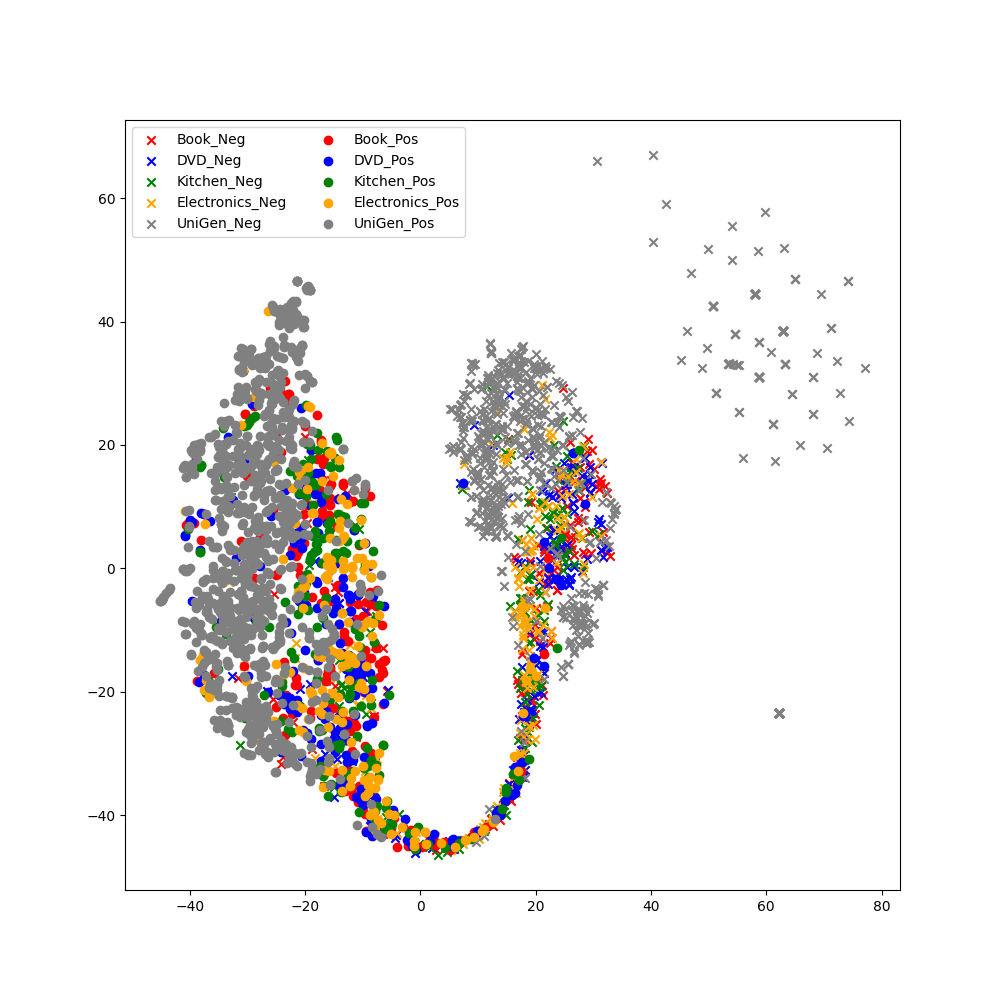}
    \caption{T-SNE visualization of the encoded representation of the RoBERTa model trained using \textsc{UniGen}. The model was trained only on the data generated using \textsc{UniGen}, which is shown in gray color. We used the test set of the multi-domain review dataset.}
\label{fig-tsne}
\end{figure}

To intuitively examine the domain generalizability of \textsc{UniGen}, we plotted the T-SNE \cite{van2008visualizing} visualization of the features interpreted by the RoBERTa-based TAM trained using \textsc{UniGen}. Figure~\ref{fig-tsne} depicts the visualization results. These results suggest that the single TAM classified the given data from every domain \textit{without} explicit training or prior information about the domains, thus demonstrating the unique efficiency of \textsc{UniGen}.

Table~\ref{tab-examples} presents examples of the sentences generated using \textsc{UniGen}. These examples showcase that \textsc{UniGen} can generate domain-invariant sentences with the designated labels. By training TAMs on these data, it is possible to distill the domain generalizability of PLMs into TAMs.

\subsection{Ablation Study}
This section describes the ablation studies conducted to offer rationales for the engineering choices made in this study. We used the DistilBERT-based TAM for these experiments.

\begin{table}[t]
\begin{center}
\resizebox{\columnwidth}{!}{
\begin{tabular}{c|cccccccc}
\Xhline{3\arrayrulewidth}
DistilBERT                                                                    & SST-2 & IMDB  & Rotten & Amazon & Yelp  & CR    & Tweet & Average        \\ \hline\hline
\begin{tabular}[c]{@{}c@{}}\textsc{UniGen}\end{tabular}                       & 77.67 & 67.81 & 73.16  & 75.06  & 74.81 & 79.86 & 81.41 & \textbf{75.68} \\
\begin{tabular}[c]{@{}c@{}}\textsc{UniGen}\\ w/ Hard Relabeling\end{tabular}  & 77.18 & 67.18 & 72.37  & 72.91  & 72.95 & 78.14 & 80.39 & 74.45          \\
\begin{tabular}[c]{@{}c@{}}\textsc{UniGen}\\ w/o Relabeling\end{tabular}      & 76.34 & 66.58 & 71.78  & 70.63  & 70.97 & 76.59 & 79.62 & 73.22          \\
\begin{tabular}[c]{@{}c@{}}\textsc{UniGen}\\ w/o Denoising MB\end{tabular}    & 77.06 & 67.13 & 72.04  & 74.69  & 73.66 & 78.47 & 80.84 & 74.84          \\
\begin{tabular}[c]{@{}c@{}}\textsc{UniGen}\\ w/o SCL\end{tabular}             & 75.53 & 66.10 & 69.63  & 71.43  & 69.58 & 77.22 & 79.31 & 72.69          \\
\begin{tabular}[c]{@{}c@{}}Combined Prompts\end{tabular}                      & 74.19 & 63.16 & 71.08  & 73.62  & 72.93 & 78.05 & 78.02 & 73.01          \\

\Xhline{3\arrayrulewidth}
\end{tabular}
}
\end{center}
\caption{Results of ablation studies on methodological choices in Section~\ref{exp:ablation1}, \ref{exp:ablation2}, and \ref{exp:ablation3}.}
\label{tab-ablation123}
\end{table}

\begin{table}[t]
\begin{center}
\resizebox{\columnwidth}{!}{
\begin{tabular}{c|cccccccc}
\Xhline{3\arrayrulewidth}
DistilBERT                                                                    & SST-2 & IMDB  & Rotten & Amazon & Yelp  & CR    & Tweet & Average        \\ \hline\hline
\begin{tabular}[c]{@{}c@{}}\textsc{UniGen}\\ w/ GPT2-XL \end{tabular}         & 77.67 & 67.81 & 73.16  & 75.06  & 74.81 & 79.86 & 81.41 & \textbf{75.68} \\
\begin{tabular}[c]{@{}c@{}}\textsc{UniGen}\\ w/ Gemma-2b\end{tabular}         & 71.50 & 69.40 & 67.04  & 76.48  & 76.89 & 77.24 & 52.03 & 70.08          \\
\begin{tabular}[c]{@{}c@{}}\textsc{UniGen}\\ w/ Qwen2-1.5B\end{tabular}       & 66.37 & 63.19 & 63.76  & 71.69  & 72.44 & 66.06 & 63.49 & 66.71          \\
\begin{tabular}[c]{@{}c@{}}\textsc{UniGen}\\ w/ Phi-1.5\end{tabular}          & 74.98 & 68.35 & 70.82  & 73.86  & 75.11 & 71.82 & 84.01 & 74.13          \\

\Xhline{3\arrayrulewidth}
\end{tabular}
}
\end{center}
\caption{Results of ablation studies on comparison between various PLMs in Section~\ref{exp:ablation4}.}
\label{tab-ablation4}
\end{table}

\subsubsection{Effectiveness of Relabeling Strategy}
\label{exp:ablation1}

First, we performed an ablation study to validate the effectiveness of the relabeling strategy discussed in Section~\ref{sec:method-relabeling}. We compared the basic approach that uses soft labels to the two other options. The first option utilizes the pseudo-relabeling process, but it assigns hard labels instead of soft labels. In other words, it only reflects the decision emanating from the PLM, not the probability. The second option completely excludes the relabeling process. While this option would generate the dataset faster than the other options, it might generate text with noisy labels, as already discussed in previous works \cite{ye2022zerogen, ye2022progen, gao2023self}.

The experimental results are presented in the second and third rows of Table~\ref{tab-ablation123}. They suggest that the use of soft labels offers practical benefits in terms of performance. This finding is consistent with that of a previous study in which the strength of soft labels was demonstrated \cite{yoo2021gpt3mix, fang2024fly}. Therefore, according to the results of this ablation study, relabeling the generated data with the assignment of soft labels is effective for mitigating the issue of noisy labels.

\subsubsection{Effectiveness of Supervised Contrastive Learning and Denoising Memory Bank}
\label{exp:ablation2}

Second, we conducted a comparison to investigate the effectiveness of supervised contrastive learning, which was discussed in Section~\ref{sec:method-supcon}, and denoising memory bank, which was discussed in Section~\ref{sec:method-denoisingMB}. The results of the comparison are presented in fourth and fifth rows of Table~\ref{tab-ablation123}. Intuitively, if the quality of each of the data in the dataset is given as a weight, it would be effective to employ only high-quality samples for comparing contrastive learning rather than utilizing all data, regardless of their quality. The experimental result in the fourth row demonstrated that the use of a denoising memory bank yielded a performance gain, which was consistent with our intuition. Similarly, the result in the fifth row suggests that supervised contrastive learning plays a crucial role in \textsc{UniGen}.

\subsubsection{Comparison with Combined Domain-specific Datasets}
\label{exp:ablation3}
Third, we compared the performance of the TAMs trained with two different synthetic datasets. The first uses the synthetic dataset generated with the prompt of \textsc{UniGen}, and the second uses the concatenation of datasets generated with five different domain-specific prompts used in the other experiments. For this experiment, we only differentiated the synthetic dataset used for training and set every other configuration identical, such as the usage of pseudo-relabeling and denoised memory bank, as well as other hyperparameters. The result of the ablation study is presented in the last row of Table~\ref{tab-ablation123}. The result indicates that the model trained with the dataset generated by the universal prompt in \textsc{UniGen} demonstrated better average performance. This suggests that the broad understanding of the label space offered by the synthetic dataset generated by \textsc{UniGen} plays an important role in domain generalization.

\subsubsection{Comparison between PLMs for Data Generation}
\label{exp:ablation4}
Lastly, we evaluated the performance of TAMs trained using various PLMs. Initially, we utilized GPT2-XL as the PLM for data generation. In this experiment, we extended the evaluation by incorporating more recent models as data generators. Specifically, we compared the performance of TAMs trained with \textsc{UniGen} using Gemma-2b \cite{team2024gemma}, Qwen2-1.5B \cite{yang2024qwen2}, and Phi-1.5 \cite{li2023textbooks}, which are more recent models with parameter sizes comparable to GPT2-XL. All other configurations, aside from the PLM used for data generation, were kept consistent with the original GPT2-XL-based TAM.

Table~\ref{tab-ablation4} presents the results of this experiment. Interestingly, the findings suggest that employing more recent PLMs does not necessarily lead to better performance in \textsc{UniGen}. The TAM trained with GPT2-XL, our original choice for data generation, achieved the highest average performance. This aligns with previous studies, which indicate that using larger PLM does not always result in superior outcomes \cite{ye2022zerogen}. However, despite using identical hyperparameters and prompts to ensure a fair comparison, it is important to recognize that optimal hyperparameters, such as top-k, top-p, and $\tau_{\textit{RE}}$, as well as the prompt configurations, may vary for each PLM. Future research could focus on developing a unified framework to optimize hyperparameters and prompts for each PLMs, akin to methods like AutoAugment \cite{cubuk2019autoaugment, ren2021text}.

\section{Conclusion}

In this study, we proposed \textsc{UniGen} in an attempt to achieve universal domain generalization. \textsc{UniGen} successfully transferred the domain generalizability of PLMs into orders-of-magnitude smaller TAMs. Moreover, human annotation was not required for \textsc{UniGen}, which significantly reduced the burden of acquiring labeled data from multiple source domains. Our relabeling method and denoising memory bank offered additional performance gains. Furthermore, our extensive experiments demonstrated that \textsc{UniGen} outperformed \textsc{Prompting}, facilitating light inference while preserving the domain generalizability of PLMs.

Although we explored an interesting framework for zero-shot, lightweight domain generalization, the performance of \textsc{UniGen} appears weaker than those of baseline models that are trained on each domain in several cases. It is desirable to achieve a higher level of performance than those of the in-domain baselines, which we will attempt in future work. To this end, the generation of small task-specific data for additional training of the TAM trained using \textsc{UniGen} is a possible approach, especially when a downstream task domain is introduced. By employing TAMs that are pre-trained using \textsc{UniGen} as a warm start, high performance could be achieved in the target domain with a small amount of task-specific data, which would reduce the total amount of data generated compared to that when individually training each TAM by using \textsc{ZeroGen} or \textsc{SunGen} from scratch. Another possible approach may involve combining \textsc{UniGen} with the concept of test-time learning \cite{jeong2023test}. Similar to the first strategy, it may generate small amounts of test domain-specific data given test-time data as in-context examples. We are committed to exploring these possible strategies, which will enhance the effectiveness of \textsc{UniGen}.

\section*{Limitations}
The primary limitation of \textsc{UniGen} is its relatively weaker in-domain performance than those of baselines that are trained with domain-specific datasets. While it is beneficial for its smaller parameter set and lower inference cost while maintaining the domain generalizability of PLMs, there exists a tradeoff between in-domain performance and efficiency, unlike \textsc{ZeroGen} and \textsc{SunGen}. Therefore, a method for further enhancing the performance of \textsc{UniGen} should be explored, as stated in the Conclusion section. A possible solution is a proper prompt designed for \textsc{UniGen} because the quality of the generated sentences is affected by prompt design. Even though we adapted an effective prompt designed in a previous work \cite{ye2022zerogen}, a more effective prompt for \textsc{UniGen} that aims to generate diverse and general expressions could exist.

\section*{Ethics Statement}
The data generated by the PLM may contain biased sentences, which may offend the readers. This can be attributed to the potential bias of PLMs \cite{liu2022quantifying}. These generated biased sentences do not reflect the views of the authors.

\section*{Acknowledgements}
This research was supported by Basic Science Research Program through the National Research Foundation of Korea(NRF) funded by the Ministry of Education(NRF-2022R1C1C1008534), and Institute for Information \& communications Technology Planning \& Evaluation (IITP) through the Korea government (MSIT) under Grant No. 2021-0-01341 (Artificial Intelligence Graduate School Program, Chung-Ang University).

\bibliography{custom}
\bibliographystyle{acl_natbib}

\appendix

\newpage
\section{Prompt for Each Domain}
\begin{table}[h]
\begin{center}
\resizebox{\columnwidth}{!}{
\begin{tabular}{c|c}
\Xhline{3\arrayrulewidth}
Domain                                                                          & Prompt                                                                        \\ \hline\hline
Movie                                                                           & The \textit{movie review} in [positive/negative] sentiment is:                \\
Products                                                                        & The \textit{product review} in [positive/negative] sentiment is:              \\
Restaurant                                                                      & The \textit{restaurant review} in [positive/negative] sentiment is:           \\
Electronics                                                                     & The \textit{electronics product review} in [positive/negative] sentiment is:  \\
Tweet                                                                           & The \textit{tweet} in [positive/negative] sentiment is:                       \\
\begin{tabular}[c]{@{}c@{}}\textsc{UniGen} \& \\ \textsc{Prompting}\end{tabular} & The \textit{text} in [positive/negative] sentiment is:                        \\ \Xhline{3\arrayrulewidth}
\end{tabular}
}
\end{center}
\caption{The prompt used for each domain in \textsc{ZeroGen} and \textsc{SunGen}, as well as the prompt used for \textsc{UniGen} and \textsc{Prompting}.}
\end{table}

\section{Implementation Detail}
For \textsc{UniGen}, we first generated 1,000k data from the 1.5B GPT2-XL model as $\mathcal{P}$ by using the prompt $\mathcal{T}_{\textit{uni}}$ ``\textit{The text in positive/negative sentiment is: }'', which is a slightly modified version of the best prompt suggested in a previous study. Top-k and top-p were set to 40 and 0.9 during the generation procedure, respectively. The soft relabeling process was performed using a $\tau_{\textit{RE}}$ of 0.1. After obtaining the soft labels of each of the generated samples, we filtered them using $T_{\textit{RE}}$ of 0.2. This required the largest value from the soft labels to be larger than the sum of the uniform distribution and $T_{\textit{RE}}$, for instance, 0.7 in binary classification with $T_{\textit{RE}}$ of 0.2. As an example, the sentence corresponding to the soft label $[0.64, 0.36]$ was discarded because it did not exceed the threshold.

After generation, we followed the bi-level optimization approach proposed in \textsc{SunGen} to cleanse the generated dataset and find the sample weights for 50 epochs. The outer learning rate was set to 5e-2, and we randomly sampled 50k data for each outer validation process. Then, we selected 200k data with high weights, which represent high-quality data, to train the TAMs. 

We used a one-layer bi-LSTM model for the LSTM-based TAM and the distilbert-base-uncased and roberta-base from Transformers \cite{wolf2020transformers} for the DistilBERT-based TAM and RoBERTa-based TAM, respectively. We trained the LSTM-based TAM for 5 epochs with the learning rate of 1e-3 by using the Adam \cite{kingma2015adam} optimizer. The DistilBERT-based TAM was trained for 3 epochs with a learning rate of 2e-5 by using the Adam optimizer. The RoBERTa-based TAM was trained for 3 epochs with a learning rate of 2e-5 by using the Adam optimizer. During the training process, $\alpha$ for supervised contrastive learning loss was set to 0.5, with a projection size of 256. The temperature $\tau_{\textit{SCL}}$ was set to 0.2, and the memory bank size $M$ was set to 64. The coefficient $m$ for updating the momentum encoder was set to 0.999, and the threshold of the denoising memory bank $T_{\textit{MB}}$ was set to 0.8. The dataset generation and training procedures were executed using on a single NVIDIA A100 40GB GPU. Please refer to attached source code for further details.\footnote{\url{https://github.com/c-juhwan/unigen}}

\section{Extensibility of Relabeling Strategy}

\begin{table}[h]
\begin{center}
\resizebox{\columnwidth}{!}{
\begin{tabular}{c|cccccccc}
\Xhline{3\arrayrulewidth}
DistilBERT                                                                     & SST-2 & IMDB  & Rotten & Amazon & Yelp  & CR    & Tweet & Average          \\ \hline\hline
\textsc{ZeroGen}                                                               & 80.06 & 69.13 & 74.73  & 73.02  & 72.77 & 73.59 & 74.83 & 74.02            \\
\begin{tabular}[c]{@{}c@{}}\textsc{ZeroGen}\\ w/ Hard Relabeling\end{tabular} & 80.72 & 69.25 & 73.98  & 73.41  & 73.18 & 73.76 & 74.91 & 74.17            \\
\begin{tabular}[c]{@{}c@{}}\textsc{ZeroGen}\\ w/ Soft Relabeling\end{tabular} & 81.79 & 70.40 & 75.32  & 73.65  & 73.31 & 74.72 & 75.14 & \textbf{74.90}   \\ \Xhline{3\arrayrulewidth}
\end{tabular}
}
\end{center}
\caption{Experimental result on the extensibility of relabeling strategy. We trained the TAM using \textsc{ZeroGen} based on the movie domain.}
\label{tab-ablation5}
\end{table}

We examined the extensibility of the relabeling strategy discussed in Section~\ref{sec:method-relabeling}. We applied two different options for relabeling, namely assigning hard labels and soft labels to \textsc{ZeroGen}. Table~\ref{tab-ablation5} summarizes the results. These results suggest that the relabeling strategy is beneficial for the performance of the TAM trained using \textsc{ZeroGen}. Therefore, filtering the generated data through the relabeling strategy is an extensive strategy for enhancing zero-shot learning methods based on dataset generation. Furthermore, the assignment of soft labels was more beneficial compared to the assignment of hard labels, which is consistent with the results of the ablation study described in Section~\ref{exp:ablation1}. We will further investigate the relabeling-based approach to enhance \textsc{ZeroGen} and \textsc{SunGen} in future works.

\section{Additional Experiment on Domain Generalizability}

To further reveal the domain generalizability of \textsc{UniGen}, we conducted an additional experiment on Amazon Review dataset \cite{ni2019justifying}. We used 5-core data for 29 domains and reported the performance of \textsc{Prompting} using GPT2-XL \cite{radford2019language} and RoBERTa-based TAM trained by \textsc{UniGen}. The result in Table~\ref{tab-amazon} demonstrates the performance of \textsc{UniGen} that is comparable with \textsc{Prompting}, with parameters less than 10\%. Additionally, this experiment showcases the universality of \textsc{UniGen}, the characteristics that distinguish \textsc{UniGen} from previous \textsc{ZeroGen} and \textsc{SunGen}. Compared to previous methods that would require 29 separately trained TAMs to conduct this experiment, \textsc{UniGen} only used one single TAM to perform the experiment, which exhibits the real-world applicability of \textsc{UniGen}.

\begin{table}[t]
\centering
\resizebox{\columnwidth}{!}{%
\begin{tabular}{c|cc}
\Xhline{3\arrayrulewidth}
Domain                      & \textsc{Prompting} & \textsc{UniGen}  \\ \hline\hline
Fashion                     & 93.29 & 91.16 \\ \hline
Beauty                      & 95.63 & 92.87 \\ \hline
Appliances                  & 68.27 & 79.10 \\ \hline
Arts, Crafts and Sewing     & 91.05 & 92.08 \\ \hline
Automotive                  & 91.07 & 88.23 \\ \hline
Books                       & 89.18 & 91.26 \\ \hline
CDs and Vinyl               & 82.44 & 86.42 \\ \hline
Cell Phones and Accessories & 90.47 & 88.65 \\ \hline
Clothing, Shoes and Jewelry & 91.83 & 90.80 \\ \hline
Digital Music               & 93.72 & 90.62 \\ \hline
Electronics                 & 88.68 & 88.34 \\ \hline
Gift Cards                  & 94.03 & 92.50 \\ \hline
Grocery and Gourmet Food    & 92.31 & 91.09 \\ \hline
Home and Kitchen            & 92.11 & 91.53 \\ \hline
Industrial and Scientific   & 91.07 & 92.34 \\ \hline
Kindle Store                & 89.49 & 92.76 \\ \hline
Luxury Beauty               & 90.03 & 91.82 \\ \hline
Magazine Subscriptions      & 85.97 & 89.64 \\ \hline
Movies and TV               & 86.39 & 88.19 \\ \hline
Musical Instruments         & 90.72 & 90.20 \\ \hline
Office Products             & 91.74 & 89.60 \\ \hline
Patio, Lawn and Garden      & 89.96 & 87.87 \\ \hline
Pet Supplies                & 90.60 & 89.91 \\ \hline
Prime Pantry                & 93.64 & 88.15 \\ \hline
Software                    & 82.55 & 83.39 \\ \hline
Sports and Outdoors         & 88.63 & 90.36 \\ \hline
Tools and Home Improvement  & 87.41 & 88.90 \\ \hline
Toys and Games              & 91.54 & 92.02 \\ \hline
Video Games                 & 85.79 & 86.07 \\ \hline\hline
\textit{Average}            & 89.30 & 89.51 \\
\Xhline{3\arrayrulewidth}
\end{tabular}
}
\caption{The result of the experiment on the Amazon Review dataset.}
\label{tab-amazon}
\end{table}

\section{Additional Study on the Performance of \textsc{UniGen} on Small-sized TAMs}
\label{app:small-tam}

\begin{table*}[t]
\begin{center}
\resizebox{\textwidth}{!}{
\begin{tabular}{cc|cc|ccccccc|c}
\Xhline{3\arrayrulewidth}
Model                        & \#Param                & Training Domain             & Setup              & SST-2                & IMDB                 & Rotten               & Amazon               & Yelp                 & CR                   & Tweet                & Average              \\
Test Domain                  &                        &                             &                    & \multicolumn{3}{c}{Movie}                                          & Products             & Restaurant           & Electronics          & Tweet                &                      \\ \hline\hline
GPT2-XL                      & 1.5B                   & -                           & \textsc{Prompting} & 82.15                & 70.26                & 77.56                & 79.06                & 78.04                & 80.30                & 80.38                & 78.25                \\ \Xhline{2\arrayrulewidth}
\multirow{11}{*}{LSTM}       & \multirow{11}{*}{7M}   & \multirow{2}{*}{Movie}      & \textsc{ZeroGen}   & 75.11                & 66.39                & 69.85                & 67.24                & 70.25                & 69.32                & 63.43                & 68.80                \\
                             &                        &                             & \textsc{SunGen}    & \textbf{78.79}       & \textbf{69.97}       & \textbf{73.76}       & 72.15                & 73.21                & 70.39                & 66.84                & \textbf{72.16}       \\ \cline{3-12} 
                             &                        & \multirow{2}{*}{Products}   & \textsc{ZeroGen}   & 64.26                & 61.82                & 60.13                & 70.32                & 67.78                & 69.46                & 62.29                & 65.15                \\
                             &                        &                             & \textsc{SunGen}    & 67.83                & 63.87                & 63.46                & \textbf{74.43}       & 73.71                & 73.35                & 63.51                & 68.59                \\ \cline{3-12} 
                             &                        & \multirow{2}{*}{Restaurant} & \textsc{ZeroGen}   & 67.41                & 63.01                & 62.74                & 68.73                & 75.51                & 69.23                & 66.35                & 63.28                \\
                             &                        &                             & \textsc{SunGen}    & 69.15                & 66.62                & 64.56                & 73.22                & {\ul \textbf{79.56}} & 70.12                & 67.43                & 70.09                \\ \cline{3-12} 
                             &                        & \multirow{2}{*}{Electronics}& \textsc{ZeroGen}   & 64.69                & 59.13                & 60.20                & 66.34                & 67.72                & 72.50                & 60.25                & 64.40                \\
                             &                        &                             & \textsc{SunGen}    & 68.38                & 64.33                & 63.25                & 72.61                & 73.01                & \textbf{76.18}       & 66.78                & 69.22                \\ \cline{3-12} 
                             &                        & \multirow{2}{*}{Tweet}      & \textsc{ZeroGen}   & 61.84                & 60.17                & 59.43                & 64.13                & 63.68                & 65.02                & 74.10                & 64.05                \\
                             &                        &                             & \textsc{SunGen}    & 66.57                & 63.96                & 64.21                & 69.36                & 71.68                & 72.57                & {\ul \textbf{81.29}} & 69.95                \\ \cline{3-12} 
                             &                        & -                           & \textsc{UniGen}    & 64.15                & 60.02                & 60.51                & 63.82                & 63.20                & 69.61                & 70.32                & 64.52                \\  \Xhline{2\arrayrulewidth}
                             
\multirow{11}{*}{CNN}        & \multirow{11}{*}{10M}  & \multirow{2}{*}{Movie}      & \textsc{ZeroGen}   & 74.34                & 67.91                & 70.22                & 68.69                & 71.03                & 70.89                & 64.77                & 69.69                \\
                             &                        &                             & \textsc{SunGen}    & \textbf{76.98}       & \textbf{68.97}       & \textbf{73.49}       & 73.04                & 73.97                & 71.55                & 69.43                & \textbf{72.49}       \\ \cline{3-12} 
                             &                        & \multirow{2}{*}{Products}   & \textsc{ZeroGen}   & 63.46                & 62.13                & 60.35                & 70.94                & 68.34                & 72.34                & 65.71                & 66.18                \\
                             &                        &                             & \textsc{SunGen}    & 65.89                & 63.27                & 61.97                & \textbf{73.98}       & 72.81                & 74.02                & 67.38                & 68.47                \\ \cline{3-12} 
                             &                        & \multirow{2}{*}{Restaurant} & \textsc{ZeroGen}   & 67.76                & 64.18                & 62.16                & 70.17                & 76.65                & 71.27                & 65.43                & 68.23                \\
                             &                        &                             & \textsc{SunGen}    & 68.86                & 65.62                & 64.96                & 73.20                & \textbf{77.87}       & 72.43                & 68.36                & 70.19                \\ \cline{3-12} 
                             &                        & \multirow{2}{*}{Electronics}& \textsc{ZeroGen}   & 65.05                & 63.04                & 62.13                & 67.19                & 69.50                & 73.66                & 63.23                & 66.26                \\
                             &                        &                             & \textsc{SunGen}    & 67.43                & 65.13                & 63.25                & 70.82                & 72.79                & \textbf{77.42}       & 67.19                & 69.15                \\ \cline{3-12} 
                             &                        & \multirow{2}{*}{Tweet}      & \textsc{ZeroGen}   & 60.56                & 60.68                & 61.33                & 64.91                & 64.37                & 66.86                & 75.62                & 64.90                \\
                             &                        &                             & \textsc{SunGen}    & 65.12                & 61.56                & 63.42                & 66.45                & 68.46                & 68.71                & \textbf{80.17}       & 67.70                \\ \cline{3-12} 
                             &                        & -                           & \textsc{UniGen}    & 62.31                & 60.48                & 61.82                & 61.08                & 61.63                & 68.24                & 65.95                & 63.07                \\ \Xhline{2\arrayrulewidth}
                             
\multirow{11}{*}{TinyBERT}   & \multirow{11}{*}{14.5M}& \multirow{2}{*}{Movie}      & \textsc{ZeroGen}   & 78.95                & 68.37                & 71.34                & 70.59                & 71.35                & 71.18                & 68.94                & 71.53                \\
                             &                        &                             & \textsc{SunGen}    & \textbf{80.78}       & \textbf{69.86}       & \textbf{73.47}       & 72.36                & 72.42                & 73.75                & 70.81                & 73.35                \\ \cline{3-12} 
                             &                        & \multirow{2}{*}{Products}   & \textsc{ZeroGen}   & 69.22                & 62.79                & 63.44                & 72.57                & 69.70                & 73.22                & 71.21                & 68.88                \\
                             &                        &                             & \textsc{SunGen}    & 71.74                & 64.38                & 64.51                & \textbf{75.81}       & 73.76                & 74.17                & 72.86                & 71.03                \\ \cline{3-12} 
                             &                        & \multirow{2}{*}{Restaurant} & \textsc{ZeroGen}   & 75.79                & 64.62                & 65.53                & 71.33                & 77.10                & 73.52                & 70.84                & 71.25                \\
                             &                        &                             & \textsc{SunGen}    & 77.45                & 67.41                & 68.01                & 74.41                & {\ul \textbf{79.16}} & 75.86                & 72.11                & 73.49                \\ \cline{3-12} 
                             &                        & \multirow{2}{*}{Electronics}& \textsc{ZeroGen}   & 71.22                & 64.37                & 63.06                & 69.51                & 70.75                & 75.71                & 69.49                & 69.16                \\
                             &                        &                             & \textsc{SunGen}    & 73.10                & 65.81                & 66.71                & 71.33                & 74.86                & 78.43                & 73.88                & 72.02                \\ \cline{3-12} 
                             &                        & \multirow{2}{*}{Tweet}      & \textsc{ZeroGen}   & 70.76                & 63.40                & 64.43                & 68.74                & 70.44                & 73.72                & 78.14                & 69.95                \\
                             &                        &                             & \textsc{SunGen}    & 73.94                & 64.87                & 66.31                & 71.39                & 72.21                & 78.16                & {\ul \textbf{81.23}} & 72.59                \\ \cline{3-12} 
                             &                        & -                           & \textsc{UniGen}    & 76.74                & 66.88                & 69.63                & 73.29                & 72.10                & \textbf{78.64}       & {\ul 80.52}          & \textbf{73.97}       \\ \Xhline{3\arrayrulewidth}
\end{tabular}
}
\end{center}
\caption{Result of ablation study that examines the performance of \textsc{UniGen} and baselines on small-sized TAMs. The performance of TAM, which is superior to that of \textsc{Prompting}, is underlined, and the best result in each test dataset within the group for each TAM is presented in boldface.}
\label{tab-small}
\end{table*}

We found that \textsc{UniGen} suffers to exhibit its performance on the LSTM model from the experiment in Table~\ref{tab-main}. To further investigate this phenomenon, we expand our experiment into two different small-sized TAMs: TextCNN \cite{kim2014convolutional} and TinyBERT \cite{jiao2020tinybert}. Table~\ref{tab-small} showcases the result of the additional experiment. In the case of TextCNN-based TAM, baseline methods such as \textsc{ZeroGen} and \textsc{SunGen} demonstrated comparable or slightly higher performance compared to that of LSTM-based TAM. Nonetheless, TextCNN-based TAM trained on \textsc{UniGen} reported slightly worse performance compared to LSTM-based TAM despite increased parameter size. We hypothesize that this phenomenon is owing to the architecture of TextCNN, which leverages CNN layers that have fixed window size, leading to limited ability to understand the context of diverse expression generated by \textsc{UniGen}. On the contrary, TinyBERT-based TAM trained on \textsc{UniGen} exhibited the best average performance among the baselines. Furthermore, its average performance is comparable to DistilBERT-based TAM despite a much smaller parameter size. It is noteworthy that TinyBERT is also a model that has a general understanding of the language through knowledge distillation from BERT. Through this investigation, we reveal that the pre-trained knowledge of the TAM aids the successful training of the TAM through \textsc{UniGen}.

\end{document}